%% file: main.tex
\documentclass[10pt,twocolumn,letterpaper]{article}

\usepackage[pagenumbers]{cvpr} 

\usepackage{graphicx}
\usepackage{amsmath}
\usepackage{amssymb}
\usepackage{booktabs}
\usepackage{multirow}
\usepackage{adjustbox}
\usepackage{wrapfig}
\usepackage{multicol}
\usepackage{lipsum}
\usepackage{stfloats}
\usepackage[accsupp]{axessibility}
\usepackage[pagebackref,breaklinks,colorlinks]{hyperref}

\usepackage[capitalize]{cleveref}
\crefname{section}{Sec.}{Secs.}
\Crefname{section}{Section}{Sections}
\Crefname{table}{Table}{Tables}
\crefname{table}{Tab.}{Tabs.}

\def\cvprPaperID{8221} 
\def\confName{CVPR}
\def\confYear{2023}

\begin{document}

\title{Improving Visual Grounding by Encouraging Consistent Gradient-based Explanations}

\author{Ziyan Yang\\
Rice University\\
{\tt\small zy47@rice.edu}
\and
Kushal Kafle\\
Adobe Research\\
{\tt\small kkafle@adobe.com}
\and
Franck Dernoncourt\\
Adobe Research\\
{\tt\small dernonco@adobe.com}
\and
Vicente Ordonez\\
Rice University\\
{\tt\small vicenteor@rice.edu}
}
\maketitle

\newcommand{\model}{\mbox{AMC-ALBEF}\xspace}
\newcommand{\objective}{AMC\xspace}
\let\vec\mathbf

\begin{abstract}
    \vspace{-0.1in}
  We propose a margin-based loss for tuning joint vision-language models so that their gradient-based explanations are consistent with region-level annotations provided by humans for relatively smaller grounding datasets. We refer to this objective as Attention Mask Consistency (\objective) and demonstrate that it produces superior visual grounding results than previous methods that rely on using vision-language models to score the outputs of object detectors. 
  Particularly, a model trained with \objective on top of standard vision-language modeling objectives obtains a state-of-the-art accuracy of $86.49\%$ in the Flickr30k visual grounding benchmark, an absolute improvement of 
  $5.38\%$ when compared to the best previous model trained under the same level of supervision. Our approach also performs exceedingly well on established benchmarks for referring expression comprehension where it obtains 80.34\% accuracy in the easy test of RefCOCO+, and 64.55\% in the difficult split. \objective is effective, easy to implement, and is general as it can be adopted by any vision-language model, and can use any type of region annotations. 
\end{abstract}


\vspace{-0.2in}
\section{Introduction}
    
Vision-language pretraining using images paired with captions has led to models that can transfer well to an array of tasks such as visual question answering, image-text retrieval and visual commonsense reasoning~\cite{lu2019vilbert,li2019visualbert,chen2020uniter}. Remarkably, some of these models are also able to perform visual grounding by relying on gradient-based explanations. 
While Vision-Language Models (VLMs) take advantage of the vast amounts of images and text that can be found on the web, carefully curated data with grounding annotations in the form of boxes, regions, or segments is considerably more limited. Our work aims to improve the grounding or localization capabilities of vision-language models further by tuning them under a training objective that encourages their gradient-based explanations to be consistent with human-provided region-based annotations from visually grounded data when those are available. 

\begin{figure}[t!]
\begin{center}
\includegraphics[width=0.48\textwidth]{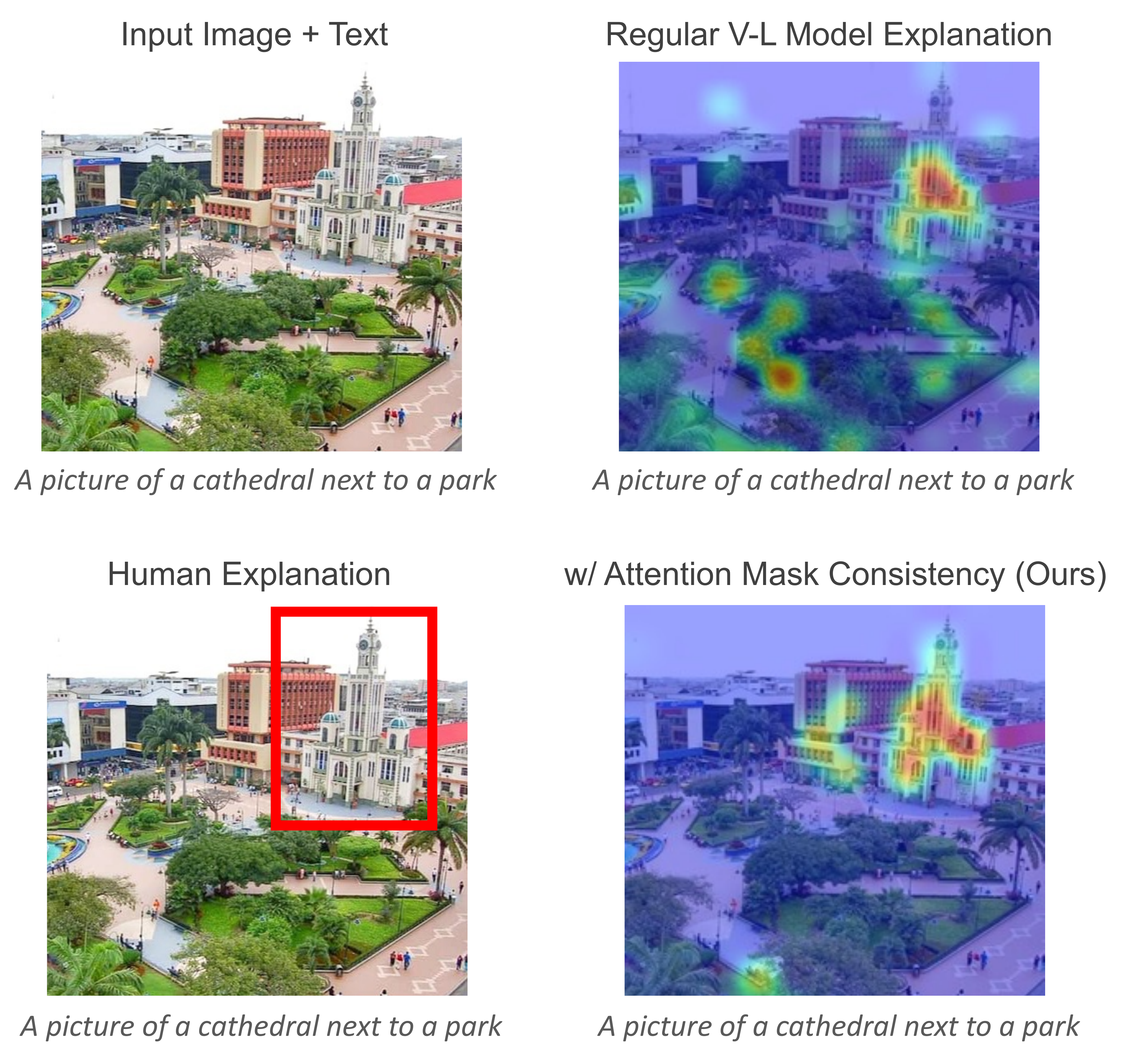}
\vspace{-0.05in}
\caption{Gradient-based methods can generate heatmaps that explain the match between images and text for a Vision-language model (VLM). Our work aims to improve their ability to produce visual groundings by directly optimizing their gradient-based explanations so that they are consistent with human annotations provided for a reduced set of images.}
\label{fig:lead}
\vspace{-0.3in}
\end{center}
\end{figure}

Vision-language transformers extend the success of masked language modeling (MLM) to multi-modal problems. In vision-language transformers, objectives such as image-text matching (ITM), and image-text contrastive losses (ITC) are used in addition to MLM to exploit commonalities between images and text~\cite{li2019visualbert,lu2019vilbert,chen2020uniter,li2021align}. We further extend these objectives to include our proposed Attention Mask Consistency (\objective) objective. Our formulation is based on the observation that gradient-based explanation maps obtained using methods such as GradCAM~\cite{selvaraju2017grad}, can be used to explain the image-text matching of a VLM. Our \objective objective explicitly optimizes these explanations during training so that they are consistent with region annotations. Figure~\ref{fig:lead} illustrates an example input image and text pair along with a gradient-based explanation obtained from a VLM model, a region annotation provided by a human, and an improved gradient-based explanation after the VLM model was tuned under our proposed objective. 

\begin{figure*}[t!]
\begin{center}
\includegraphics[width=0.84\textwidth]{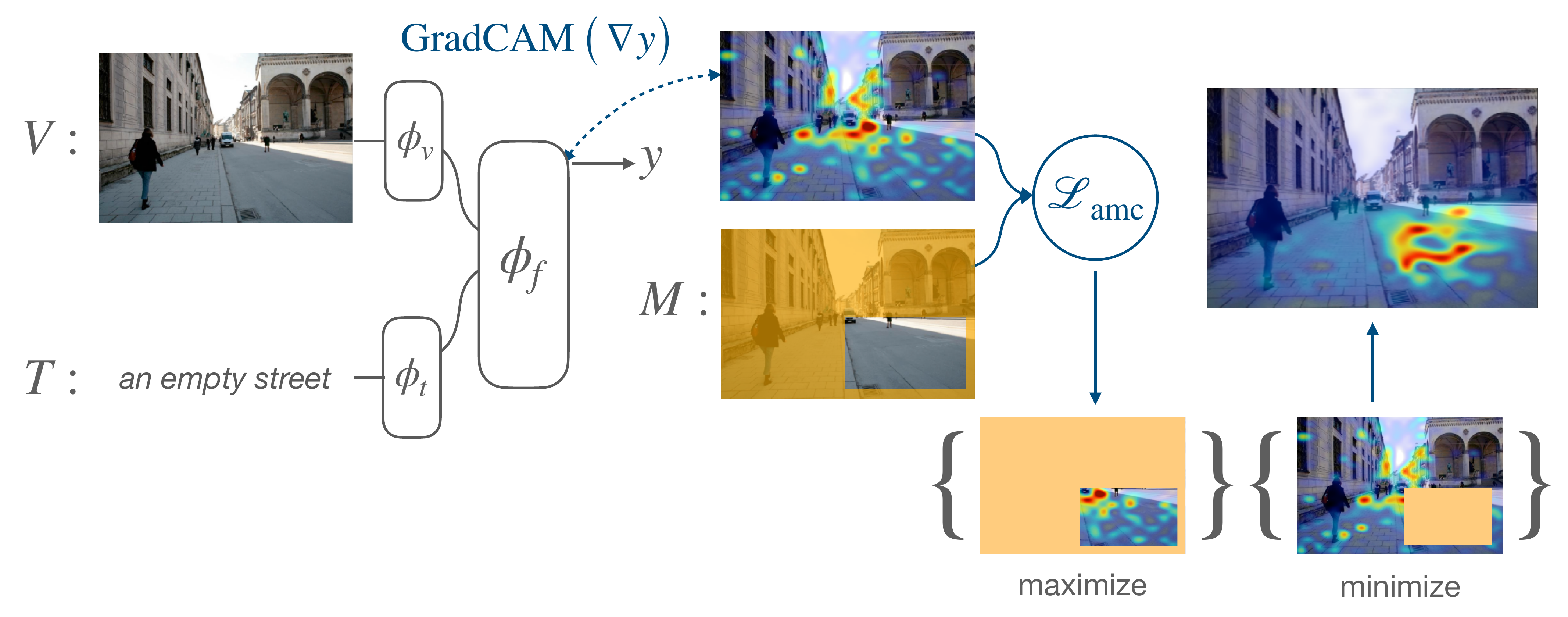}
\vspace{-0.15in}
\caption{Overview of our method. Among other objectives, standard vision-language models are trained to produce a matching score $y$ given an input image-text pair $(V,T)$. For inputs containing an extra level of supervision in the form of region annotations (e.g.~a triplet $(V,T,M)$), where M is a binary mask indicating the regions annotated by a human, we optimize the GradCAM~\cite{selvaraju2017grad} gradient-based explanations of the model so that the produced explanations are consistent with region annotations using $\mathcal{L}_{\text{amc}}$ by maximizing the energy in the heatmap that falls inside the region annotation and minimizing what falls outside. We accomplish this through soft margin losses as described in Sec.~\ref{subsec:amc}.
}
\label{fig:model}
\vspace{-0.2in}
\end{center}
\end{figure*}

Our work builds particularly upon the ALBEF model~\cite{li2021align} which incorporates a vision-language model architecture based on transformers~\cite{vaswani2017attention} and has already demonstrated off-the-shelf grounding capabilities using GradCAM. Gradient-based explanations in the form of heatmaps have been used extensively to explain the areas of the input images that most impact an output value of a model. In our formulation we actively leverage these heatmaps by designing a loss function that encourages most of the energy in the heatmaps to fall within the areas of the input image that most align with human provided region annotations. Figure~\ref{fig:model} shows a detailed overview of our method and objective function. Given an input image and text pair, our goal is to maximize a soft margin between the energy of the heatmap inside the region annotation and the energy of the heatmap outside the region annotation. A soft-margin is important since typical human region annotations in the form of boxes do not exactly outline objects of different shapes, and in many cases models should still be able to ground an input text with multiple regions across the image.

We compare \objective extensively against other methods that use the same level of supervision but instead use an object detector such as Faster-RCNN~\cite{li2021align,gupta2020contrastive,lu202012,dou2021improving}. 
Our method obtains state-of-the-art {\it pointing game accuracy} on both Flickr30k and RefCOCO+. 
Our contributions can be summarized as follows: (1) We introduce a new training objective, \objective, which is effective, simple to implement and can handle multiple types of region annotations, (2) We show that \objective can improve the grounding capabilities of an existing vision-language model -- ALBEF, and (3) the resulting model is state-of-the-art in two benchmarks for phrase grounding and referring expression comprehension.


\section{Related Work}
\label{sec:related}
\noindent{\bf Vision-Language Representation Learning.} Followed by the success of pretraining methods in NLP such as BERT~\cite{devlin2018bert}, many transformer-based image-text models have been proposed to leverage benefits of pretraining on large-scale unlabeled image-text pairs \cite{li2019visualbert,li2020oscar,huang2020pixel}. While earlier pretraining methods rely on an object detector to divide an image into input tokens, some recent works, such as ALBEF \cite{li2021align}, use an end-to-end vision transformer~\cite{dosovitskiy2020image}. These pretrained models can then be finetuned to obtain impressive performance in a wide variety of vision-language tasks, such as image-text retrieval, visual question answering, and visual commonsense reasoning. While these models can perform some visual grounding by running them on the outputs of an object detector or using gradient-based explanations, they are not trained to take advantage of grounded data. Our \objective objective provides this additional capability by leveraging gradient-based explanations that can be easily obtained for a large variety deep learning models.  

\vspace{0.05in}
\noindent{\bf Gradient-based Localization.} Localizing the most discriminative areas of an image for a given task has been widely used as a tool to provide visual explanation about a model. Class activation maps~(CAM)~\cite{zhou2016learning}  were proposed to provide weighted feature maps for any networks with minimal modifications to the model. Gradient-weighted Class Activation Mapping ($\mathrm{GradCAM}$) \cite{selvaraju2017grad} improves CAM by directly using gradients to obtain weighted feature maps without the need for model modifications or retraining; 
The attention maps generated by these methods can be directly optimize to guide the model toward solutions that are more consistent with human-based annotations.
Our proposed method is also based on GradCAM heatmaps. However, we use GradCAM during training to guide the generated heatmaps to achieve better consistency with known region and phrases that describe them. Recently, Pham~et~al~\cite{pham2021learning} explored a similar idea by using segmentation masks to guide attention maps to focus on significant image regions for an attribute prediction task. Selvaraju~et~al~\cite{selvaraju2021casting} use saliency maps generated using Deep-USPS~\cite{nguyen2019deepusps} at training time to guide attention maps in order to improve self-supervised representation learning.
Similarly Pillai~et~al~\cite{pillai2022consistent} rely on consistent explanations for generic representation learning using contrastive objectives. Our goal in using supervision on top of gradient-based heatmaps is to directly leverage these heatmaps to evaluate on visual grounding. 

\vspace{0.05in}
\noindent{\bf Visual Grounding Methods.} Visual Grounding is a task that requires a model to select the region of an image described by a phrase. 
Several methods have been proposed to ground phrases to regions of an image, typically a bounding box~\cite{li2021align,gupta2020contrastive,lu202012,dou2021improving}. Visual grounding has also been used to improve performance on downstream tasks such as VQA~\cite{selvaraju2019taking}. These methods take advantage of object detectors which can provide high quality locations. The recently proposed GLIP model~\cite{li2022grounded} incorporates an object detection model as part of its grounding objective, effectively combining vision-language pretraining with bounding box localization. 
Our work instead of outputting a box, optimizes its own gradient-based model explanations. Since our model does not output bounding boxes but heatmaps as an output it can generate more general groundings for phrases or objects that can not be mapped to a box such as stuff categories or references to multiple objects. Moreover, \objective can be used to improve an existing vision-language model such as ALBEF~\cite{li2021align} without retraining from scratch. As vision-language models become larger and more robust, our proposed \objective objective can be readily applied.

\section{Method}
\label{sec:method}

Vision-language pretraining consists of exploiting the structure of each input modality as well as their interactions. Our base model consists of three transformer encoders~\cite{vaswani2017attention,devlin2018bert}: An image encoder $\phi_v$, a text encoder $\phi_t$, and a multimodal fusion encoder $\phi_f$. An input image $V$ is encoded into a sequence of visual tokens $\{\Vec{v}_{\mathrm{cls}},\Vec{v}_1,\Vec{v}_2,...,\Vec{v}_n\}$ and the text encoder encodes the input text $T$ as a sequence of tokens $\{\Vec{t}_{\mathrm{cls}},\Vec{t}_1,\Vec{t}_2,...,\Vec{t}_m\}$, where $\Vec{v}_{\mathrm{cls}}$ and $\Vec{t}_{\mathrm{cls}}$ are the embeddings of the \texttt{[CLS]} token for each transformer respectively. For each image-text pair drawn from a dataset $(V, T)\sim D$, a binary variable $y$ represents whether the pair correspond with each other, i.e,~whether the text actually describes the paired image. However, for some images a triplet $(V,T,M)\sim D$ might be available, where $M$ additionally contains a region annotation, in the form of a binary mask, indicating the part of input image $V$ that text $T$ describes. In the following section we describe standard objectives used to capture intra-modality and inter-modality structure (Sec.~\ref{subsec:objectives}), and then we describe our attention mask consistency objective (Sec.~\ref{subsec:amc}).

\subsection{Standard Model Training Objectives}
\label{subsec:objectives}

\paragraph{Masking Language Modeling (MLM)} Originally introduced by BERT~\cite{devlin2018bert} in the context of language transformers, this objective has been adapted to multiple vision-language pretraining models such as~\cite{li2019visualbert,lu2019vilbert,chen2020uniter} and is inspired by a long history in NLP of exploiting distributional semantics. The goal is to capture structure in the text by forcing the model to infer missing words from the input text. Each token in the input text is masked randomly with a small probability (usually $15\%$) and the model is then optimized to recover the masked tokens using information from both the remaining input text and the input image. Assume an input masked text is represented by $T^{-m}$, and the masked token is represented as a one-hot vector $\Vec{t}^{m}$, the objective will be expressed as:
\begin{equation}
\mathcal{L}_{\mathrm{mlm}} = \;\mathbb{E}_{(V,T^{-m})\sim D}\; \mathrm{H}\left(\Vec{t}^m, \phi_f^m\left(\phi_v(V), \phi_t(T^{-m})\right)\right),
\end{equation}
where $\mathrm{H}(\cdot,\cdot)$ is the cross-entropy between the missing token $\Vec{t}^m$ and a probability distribution over tokens output by a function $\phi_f^m$ which augments $\phi_f$ with a linear projection layer and softmax function over a corresponding output embedding. This objective is optimized over a large sample of choices for masked tokens and image-text pairs.

\paragraph{Image Text Matching (ITM)} Another common objective inspired by BERT's next sentence prediction objective, consists of image text matching. The purpose of this loss is to push the model to learn if a text and an image are matched. The output of the \texttt{[CLS]} token will be used to generate the output for this objective by adding a linear layer and a softmax activation function. We denote this entire operation as $\phi_f^{\mathrm{cls}}$. The objective is therefore defined as follows: 
\begin{equation}
\mathcal{L}_{\mathrm{itm}} = \;\mathbb{E}_{(V,T)\sim D}\;\mathrm{H}\left(\Vec{y}, \phi_f^{\mathrm{cls}}\left(\phi_v\left(V\right), \phi_t\left(T\right)\right)\right),
\end{equation}
where $\Vec{y}$ is a one-hot vector with two entries $\left[y, 1-y\right]$ indicating whether the drawn sample $(V,T)$ corresponds to a matching image-text pair or not.


\paragraph{Image-Text Contrastive Loss (ITC)} This objective has been useful in weakly supervised grounding~\cite{gupta2020contrastive,li2021align}. We follow ALBEF because it uses momentum distillation to potentially leverage a larger amount of negative image-text pairs. Assuming that each image-text pair is considered within a sample batch of $K$ image-text pairs, this loss is defined as follows:
\begin{equation}
\begin{split}
\mathcal{L}_{\mathrm{itc}} = \mathbb{E}_{(V,T)\sim D}\;\frac{1}{2}\biggl[
&\mathrm{H}\biggl(\Vec{y}, s(V, T)/\sum_{k=1}^{K}s(V, T_k)\biggr)+ \\
&\mathrm{H}\biggl(\Vec{y}, s(T, V)/\sum_{k=1}^{K}s(T, V_k)\biggr)\biggr],
\end{split}
\end{equation}
where $s(V,T) = \exp(\phi_v(V) \cdot \phi_t(T)/\tau)$ computes a score between the output $\texttt{[CLS]}$ token representations for the encoder transformer of each modality and $\tau$ is a temperature parameter, and $s(T,V)$ is defined similarly. The goal of this loss is to push for matching image-text pairs to have a closer representation than any non-matching image-text pair.

\subsection{Attention Map Consistency (AMC)}
\label{subsec:amc}
In this section we explain in detail our proposed attention map consistency loss. 
Our proposed loss relies on first producing explanation heatmaps or ``attention maps'' using the $\mathrm{GradCAM}$ method~\cite{selvaraju2017grad}. In the context of vision-language transformers, this method can be used to highlight regions in the image that contribute to an image matching to an arbitrary input text, e.g., given an input image such as the one in Fig.~\ref{fig:model}, and an input text such as \emph{an empty street}, we can generate a $\mathrm{GradCAM}$ visualization of areas in the input image that contribute to their matching score using $\mathcal{L}_{\mathrm{itm}}$.


We assume that for a subset of images in our dataset we can obtain a triplet $(V,T,M)$ where $M \in \{0,1\}^2$ is a binary mask such that $M_{i,j}$ is $1$ if the location $i,j$ is inside region or $0$ otherwise, $V$ is the input image, and $T$ is an input text describing region $M$. This assumption is generally fair in comparison to previous works that instead leverage images annotated with labels and bounding boxes to train an object detector. In our case, we can easily support this setup by turning a label annotation, e.g.,~\texttt{dog} into a region textual caption by prompt engineering, e.g.,~\emph{an image of a dog}. However, our binary masks are not restricted to being boxes.

In order to compute a $\mathrm{GradCAM}$ heatmap, we first extract an intermediate feature map $F_z$ in the multimodal fusion transformer $\phi_f$ and denote this function as $\phi_z$:
\begin{equation}
F_z = \phi_z\left(\phi_v(V),\; \phi_t(T)\right).
\end{equation}
Then, we calculate the gradient of $F_z$ with respect to the matching loss $\mathcal{L}_{\mathrm{itm}}$ of this individual sample:
\begin{equation}
G_z = \nabla \mathrm{H}\left(\Vec{y}, \phi_f^{\mathrm{cls}}\left(\phi_v\left(V\right), \phi_t\left(T\right)\right)\right).
\end{equation}
Next, we calculate a $\mathrm{GradCAM}$ attention heatmap $A$ using $F_z$ and $G_z$ as follows:
\begin{equation}
A = \mathrm{ReLU}(F_z \odot G_z),
\end{equation}
where $\odot$ is an element-wise multiplication. This heatmap is resized to the resolution of input images, and identifies which area in the image explains the model decision for its matching score. 

The next step is to leverage the region annotations $M$ so that the model focuses its heatmap scores in $A$ inside the region of interest indicated by $M$. We first propose $\mathcal{L}_{\mathrm{mean}}$ where we optimize a max margin loss so that the mean value of the heatmap inside of the region of interest is larger than the mean value of the heatmap outside as follows:
\begin{equation}
\begin{split}
\mathcal{L}_{\mathrm{mean}} = \quad\quad\quad\quad\quad\quad\quad\quad& \\ 
\mathbb{E}_{(V,T,M)\sim D}\; \biggl[\max(&0, \;\frac{1}{N^{c}} \sum_{i,j}\left(1-M_{i,j}\right) A_{i,j} \\
&- \frac{1}{N}\sum_{i,j}M_{i,j} A_{i,j} + \Delta_1)\biggr],
\label{eq:amc1}
\end{split}
\end{equation}
where $\Delta_1$ is a margin term, and $N=\sum_{i,j} M_{i,j}$ is the number of locations inside the region of interest and $N^c$ is the number of locations outside i.e.~$\sum_{i,j} (1 - M_{i,j})$. This loss aims to ensure that the attention map $A$ contains most of the scores inside the region $M$ subject to this margin. We also propose to jointly maximize the margin between the largest score inside the region of interest $M$ and the largest score outside the region of interest by a margin $\Delta_2$ as follows:
\begin{equation}
\begin{split}
\mathcal{L}_{\mathrm{max}} =\;
\mathbb{E}_{(V,T,M)\sim D}\; \biggl[\max(&0, \;\max_{i,j} \left(\left(1 -M_{i,j}\right) A_{i,j}\right) \\
&- \max_{i,j} \left( M_{i,j}A_{i,j} \right) + \Delta_2)\biggr].
\label{eq:amc2}
\end{split}
\end{equation}
Finally, we combine these two objectives: 
\begin{equation}
\mathcal{L}_{\mathrm{amc}} = \lambda_1 \cdot \mathcal{L}_{\mathrm{mean}} +  \lambda_2 \cdot \mathcal{L}_{\mathrm{max}},
\label{eq:amc}
\end{equation}
where $\lambda_1$ and $\lambda_2$ are empirically determined weighting coefficients. We demonstrate in our experimental section that this objective effectively encourages model explanations that provide better grounding support for tasks such as referring expression comprehension and visual grounding.

\section{Experiments}
In this section, we describe our training setup and experimental evaluations. Our evaluations revolve around tasks that require pointing to the location in an image that is referred by an input text.

\label{sec:experiments}
\subsection{Training Details}
\label{data}

Our model follows the architecture and training objectives of the ALBEF model~\cite{li2021align} which uses the ALBEF-14M dataset as source of pretraining. {ALBEF-14M} is a large image-text data collection including the following datasets: COCO~\cite{lin2014microsoft}, Visual Genome (VG) ~\cite{krishna2017visual} (excluding box annotations), SBU~\cite{ordonez2011im2text}, CC3M~\cite{sharma2018conceptual} and CC12M~\cite{changpinyo2021conceptual}. In this data collection, each image is paired with one or several image descriptions so that we can sample pairs $(V,T)\sim D$. Additionally, several vision-language transformer models such as UNITER~\cite{chen2020uniter} or VisualBERT~\cite{li2019visualbert} further leverage box annotations from the Visual Genome dataset. We use this dataset as an additional source of triplets $(V,T,M)\sim D$. We start with the ALBEF model and further finetune it for $20$ more epochs on Visual Genome with boxes using our proposed \objective loss. Next, we describe in detail how we leverage the Visual Genome dataset to produce triplets $(V,T,M)$ in more detail.

 First, we provide a more detailed description of the Visual Genome dataset. This dataset consists of $108,077$ images and annotations in multiple formats such as boxes + region descriptions, boxes + object labels, and boxes + object attributes. At a first level, annotators of this dataset were asked to provide text that describes a region of the image and to provide a bounding box that covers the region. For instance, {\em a brown dog playing with a ball}. Then, the region descriptions were shown to other annotators that were asked to select objects from these regions and provide tight bounding boxes for selected objects and attributes e.g. {\em brown dog} and {\em ball}. Region bounding boxes and object+attribute bounding boxes are different for this dataset. The object detector trained by Anderson~et~al~\cite{anderson2018bottom} on the object bounding boxes and object attributes of Visual Genome has been used by several previous visual grounding models. 
 In order to compare fairly to these methods, we develop a model using the same training split as~\cite{anderson2018bottom} and conduct experiments without the use of region descriptions. For completeness, we also conduct experiments using both boxes with attributes and boxes with region descriptions.

  \begin{table}[t]
\small
\centering
\setlength\tabcolsep{1.5pt}
\renewcommand{\arraystretch}{1.3}
\begin{tabular}{ l c c c c c cc}
\toprule
\multirow{2}{*}{\bf Method} && \multirow{2}{*}{\bf Detector}
 && \multirow{2}{*}{\bf Flickr30k} && \multicolumn{2}{c}{\bf RefCOCO+} \\
 \cmidrule{7-8}
 && && && test A & test B\\
\midrule
Align2Ground~\cite{datta2019align2ground} && {\footnotesize Faster-RCNN (VG)} && 71.00 && - & -\\
12-in-1~\cite{lu202012} && {\footnotesize Faster-RCNN (VG)} && 76.40 && - & -\\
InfoGround~\cite{gupta2020contrastive} && {\footnotesize Faster-RCNN (VG)} && 76.74 && 39.80 & 41.11\\
VMRM~\cite{dou2021improving} && {\footnotesize Faster-RCNN (VG)} && 81.11 && 58.87 & 50.32\\
\midrule
AMC$*$   && -- &&  86.49 && 78.89 & 61.16\\ 
AMC (ours) && -- && {\bf 86.59} && {\bf 80.34} & {\bf 64.55}\\ 

\bottomrule
\end{tabular}
\caption{Visual Grounding results using {\em pointing game} accuracy against the state-of-the-art for methods. For fairer comparisons, AMC$*$ indicates a version of our model restricted to using only the box and label annotations from Visual Genome (VG) that were used to train the Faster-RCNN network used in the other methods.}
\label{detectorbased}
\end{table}

 \captionsetup[table]{skip=5pt}

We construct textual descriptions for object bounding boxes using prompt engineering templates. For example, if an image contains an object {\em dog} with an attribute {\em brown}, we construct the description as {\em a brown dog}. We filter out bounding boxes smaller than $8\%$ of the whole image. 
To further increase the localization capabilities of our method, we generate prompts with spatial references. For images with objects that correspond to more than one box, we select the leftmost/rightmost, top/bottom boxes and assign more detailed prompts such as {\em [obj] on the left}, {\em [obj] on the right}, {\em top [obj]} and {\em bottom [obj]}. Moreover, if the box falls into a corner of the image, we further assign them another level of spatial information such as {\em top left}, {\em top right}, {\em bottom left} and {\em bottom right}.

We conduct experiments on single node with 8 NVIDIA A40 GPUs.
All experiments use a batch size of $512$ and a learning rate of 1e-5 with an Adam optimizer \cite{kingma2014adam}. We determine empirically based on a small validation set two margin losses: $\Delta_1=0.1$ and $\Delta_2=0.5$ and determine our weighting coefficient for our losses as $\lambda_1=0.2$ and $\lambda_2=0.8$, respectively. For data augmentation, we resize images into a resolution of $256\times256$ and apply horizontal flips, color jittering and random grayscale conversions. Our
code and data are publicly available\footnote{ \url{https://github.com/uvavision/AMC-grounding} }.

\subsection{Visual Grounding}
\label{sec:results-grounding}
Visual grounding consists in automatically associating an area of an image with an arbitrary piece of input text. A popular benchmark for this task is Flickr30k~\cite{plummer2015flickr30k}. 
We only use the validation and testing splits for this dataset and do not use it for training. Each split includes a thousand images and is used for all of our model selections and evaluations. In Flickr30k Entities, each object phrase may pair with multiple ground truth boxes in the image. Our model will take the phrase and whole image as inputs, and find the most related regions corresponding to the phrase. 

We report experimental results for Flickr30k Entities~\cite{plummer2015flickr30k} with both detector-based and detector-free methods. {\em Pointing game} accuracy is a widely used metric in previous works for this task~\cite{gupta2020contrastive,arbelle2021detector,wang2021improving,akbari2019multi}, and we follow the same setting as in~\cite{akbari2019multi} to calculate this measure: After obtaining a heatmap given an input phrase and an image, we extract the position of the maximal point of this heatmap, and if this point falls in the target box, we count this result as positive. For detector-based methods, we follow \cite{gupta2020contrastive} to calculate the {\em pointing game} accuracy by first ranking proposals generated by an object detector and then retaining one box proposal with highest score as the result. If the center point of the selected box proposal falls within the target box, this result is counted as positive.

\begin{table}[t]
\small
\centering
\setlength\tabcolsep{1.1pt}
\renewcommand{\arraystretch}{1.2}
\begin{tabular}{ l c c c c}
\toprule
\multirow{2}{*}{\bf Method} & \multirow{2}{*}{\bf VG-Boxes}  & \multirow{2}{*}{\bf Backbone} &&  \multirow{2}{*}{\bf Flickr30k}\\
&&  & \\
\midrule
gALBEF~\cite{li2021align}& no & ALBEF && 79.14 \\
\midrule 
GbS~\cite{arbelle2021detector}& no & PNASNet && 73.39 \\
MG~\cite{akbari2019multi}& no & ELMo + PNASNet && 67.60\\
GAE~\cite{chefer2021generic}& no & CLIP && 72.47\\
WWbL~\cite{shaharabany2022looking}& no & CLIP + VGG && 75.63\\
\midrule
GbS+IG~\cite{arbelle2021detector} & yes & PNASNet && 83.40 \\
GbS+12-in-1~\cite{arbelle2021detector}& yes & PNASNet && 85.90\\
AMC (ours)& yes & ALBEF && {\bf 86.59}\\

\bottomrule
\end{tabular}
\caption{Visual Grounding results using {\em pointing game} accuracy against methods that do not use object detectors or Visual Genome box supervision, showing that box supervision still makes a significant difference on this benchmark despite the fact that CLIP uses hundreds of millions of extra images for training compared to the ALBEF backbone.}
\label{detectorfree}
\end{table}

 For Align2Ground~\cite{datta2019align2ground} and 12-in-1~\cite{lu202012}, we directly show the results reported in~\cite{arbelle2021detector}. For InfoGround~\cite{gupta2020contrastive} we use their provided trained models. For VMRM~\cite{dou2021improving}, since they do not provide their trained model, we re-train it using the official code and their used features and boxes from MMF \cite{singh2020mmf} and MAF \cite{wang2020maf}. Align2Ground, 12-in-1 and VMRM all use image features generated by object detectors trained on VG boxes and attributes~\cite{anderson2018bottom}. InfoGround uses image features extracted from an object detector trained on VG boxes. We also compare to methods that do not use any form of box supervision including our backbone ALBEF as a baseline. We refer as gALBEF to our baseline which only uses GradCAM in combination with ALBEF as described in~\cite{li2021align}.
We additionally compare to MG~\cite{akbari2019multi}, GbS~\cite{arbelle2021detector} which report results on Flickr30k. Results for GAE~\cite{chefer2021generic} as well as WWbL are taken directly from~\cite{shaharabany2022looking}. In addition to fairly compare with GbS, we additionally report their results when ensembled with detector-based methods InfoGround and 12-in-1. 

Our main comparison results for methods relying on Visual Genome boxes are summarized in Table~\ref{detectorbased} and results comparing against methods that are weakly supervised and hence do not use box information are shown in Table~\ref{detectorfree}.

\subsection{Referring Expression Resolution}
\label{sec:results-grounding}

Referring expressions are textual descriptions that refer unambiguously to an object or region of an image. Users are explicitly prompted to write a textual description to refer to a specific object. However the setup is similar to the visual grounding setup and as such, many previous methods compare their results across both benchmarks. We adopt the same {\em pointing game} accuracy metric and compare our results against previous methods in two benchmark datasets: RefCOCO+~\cite{yu2016modeling,kazemzadeh2014referitgame} and ReferIt~\cite{kazemzadeh2014referitgame}. We compare against the same set of methods as in the visual grounding task except for Align2Ground~\cite{datta2019align2ground} and 12-in-1~\cite{lu202012} which do not provide results for RefCOCO+. Additionally, InfoGround~\cite{gupta2020contrastive} does not report results for RefCOCO+, therefore, we use their provided bounding boxes for COCO images~\cite{lin2014microsoft} to perform this evaluation.

  \begin{table}[b!]
\small
\centering
\setlength\tabcolsep{2.4pt}
\renewcommand{\arraystretch}{1.45}
\begin{tabular}{ l c c c c cc }
\toprule
\multirow{2}{*}{\bf Method} && \multirow{2}{*}{\bf Boxes}  &&& \multicolumn{2}{c}{\bf RefCOCO+} \\
\cmidrule{6-7}
 && & && test A & test B\\
\midrule
VMRM~\cite{dou2021improving} && FasterRCNN &  && 30.04 & 30.78\\
VMRM~\cite{dou2021improving} && MaskRCNN &  && 46.63 & 40.52\\
gALBEF~\cite{li2021align} && MaskRCNN &  && 61.70 & 42.83\\
AMC && MaskRCNN &  && {\bf 68.04} & {\bf 46.55}\\
\bottomrule
\end{tabular}
\caption{We show recall@1 results on the RefCOCO+ validation and testing sets to complement our results using pointing game accuracy.}
\label{supp:recall}
\end{table}

We describe in more detail each benchmark. RefCOCO+~\cite{yu2016modeling} is a widely used referring expression dataset including 20K images from the COCO dataset~\cite{lin2014microsoft}. 
The expressions in RefCOCO+ were collected so that they do not allow words such as {\em left} or {\em right}, making it slightly more challenging. From this dataset, we only use its validation and testing splits. The testing split of this dataset is further divided into two subsets: \texttt{test A} and \texttt{test B}, in which the former only includes people as the target objects and the latter includes all objects. The total number of testing images is $1.5\mathrm{K}$.
Results for referring expression comprehension on RefCOCO+ are included in Table~\ref{detectorbased}.

\begin{table*}
\small
\centering
\setlength\tabcolsep{2.2pt}
\renewcommand{\arraystretch}{1.12}
\begin{adjustbox}{center}
\begin{tabular}{ l c c c c c c c c c c c }
\toprule
{\bf Method} && {\bf Overall} && {\bf People} & {\bf Animals} &  {\bf Vehicles} & {\bf Instrum.} & {\bf Bodyparts} & {\bf Clothing} & {\bf Scene} & {\bf Other}\\
\midrule
MG~\cite{akbari2019multi} && 69.2 && 75.6 & 87.6 & 83.8 & 57.5 & 44.9 & 58.3& 68.2 & 59.8\\
GbS~\cite{arbelle2021detector} && 74.5 && 83.6 & 89.3 & 92.1 & 83.3 & 53.2 & 50.1 & 71.3 & 66.7\\
gALBEF~\cite{li2021align} && 79.1 && 80.1 & 89.8 & 89.8 & 83.3 & 63.3 & 85.5 & 83.8 & 70.2\\
\midrule

Align2Ground~\cite{datta2019align2ground} && 71.0 && - & - & - & - & - & - & - & -\\
12-in-1~\cite{lu202012} && 76.4 && 85.7 & 82.7 & \textbf{95.5} & 77.4 & 33.3 & 54.6 & 80.7 & 70.6\\
InfoGround~\cite{gupta2020contrastive} && 76.7 && 83.2 & 89.7 & 87.0 & 69.7 & 45.1 & 74.5 & 80.6 & 67.3\\
VMRM~\cite{dou2021improving} && 81.1 && 88.0 & 92.3 & 94.3 & 66.7 & 55.1 & 79.8 & 85.1 & 69.9\\
\midrule
AMC && \textbf{86.6} && \textbf{89.7} & \textbf{95.2} & 93.8 & \textbf{86.4} & \textbf{69.8} & \textbf{89.0} & \textbf{91.4}& \textbf{77.7}\\

\bottomrule
\end{tabular}
\end{adjustbox}
\caption{Breakdown of results by category for {\it pointing game accuracy} on Flickr30K entities visual grounding.}
\label{both}
\end{table*}

  \subsection{Box Recall Evaluation}
  
  Pointing game accuracy has been previously used for both detector-based~\cite{gupta2020contrastive} and detector-free~\cite{arbelle2021detector,akbari2019multi} methods. However, another metric that can be considered is {\em Recall@k} from detector-based methods~\cite{gupta2020contrastive,dou2021improving}. For {\em Recall@k}, a model will rank all the box proposals generated by an object detector, and select the top-k boxes as results. If a selected box and the ground truth box have an intersection over union (IoU) $\ge$ 0.5, then the selected box will be counted as positive. 
Table~\ref{supp:recall} shows results when we evaluate our method by using it to choose boxes from different bounding box proposals methods by selecting the boxes with high attention heatmap scores. 
We use boxes generated by the FasterRCNN~\cite{ren2015faster} from Gupta~et~al~\cite{gupta2020contrastive} and the MaskRCNN~\cite{he2017mask} from Yu~et~al~\cite{yu2018mattnet}. Using the MaskRCNN proposals, our method obtains consistently better results than VMRM, which is the current stats-of-the-art. However, we find this metric is influenced by the quality of boxes. For example,  using the MaskRCNN proposals will get much better results than using the FasterRCNN proposals for VMRM~\cite{dou2021improving}.

  \subsection{Discussion of Results}

Table~\ref{detectorbased} contains the main results of our paper when compared to several other methods that rely on vision-language models coupled with box-level supervision from Visual Genome through an object detector -- FasterRCNN trained on Visual Genome. Our results show a large advantage on all benchmarks but especially on RefCOCO+. We report more fine-grained results in Table~\ref{both} for Flickr30K Entities. There are eight categories in this dataset. We evaluate on each category and report the pointing game accuracy for them separately. For MG and Gbs, we report results when trained on COCO because their models achieved their best performances on Flickr30k Entities under this setting. In general, our method obtains better results for almost all the categories. For the category \texttt{vehicle} our method obtains $93.8\%$, which is only $1.7\%$ lower than the best result from the best method (12-in-1). 

In Table~\ref{detectorfree} we observe that methods that use box supervision still exhibit considerable better performance on visual grounding on Flickr30k Entities. Our method obtains $86.59\%$, which is $10.6\%$ higher than WWbL, which is the best method that does not use box supervision from Visual Genome -- however in terms of number of training images it relies on CLIP which was trained on 400M images with text compared to our method which uses 14M images with text plus 100k images with boxes. In addition we compare to GbS~\cite{arbelle2021detector} when ensembled with detector-based methods InfoGround and 12-in-1. Our method is still superior when compared to these two strong baselines. 

\begin{figure*}[t]
\begin{center}
\includegraphics[width=0.86\textwidth]{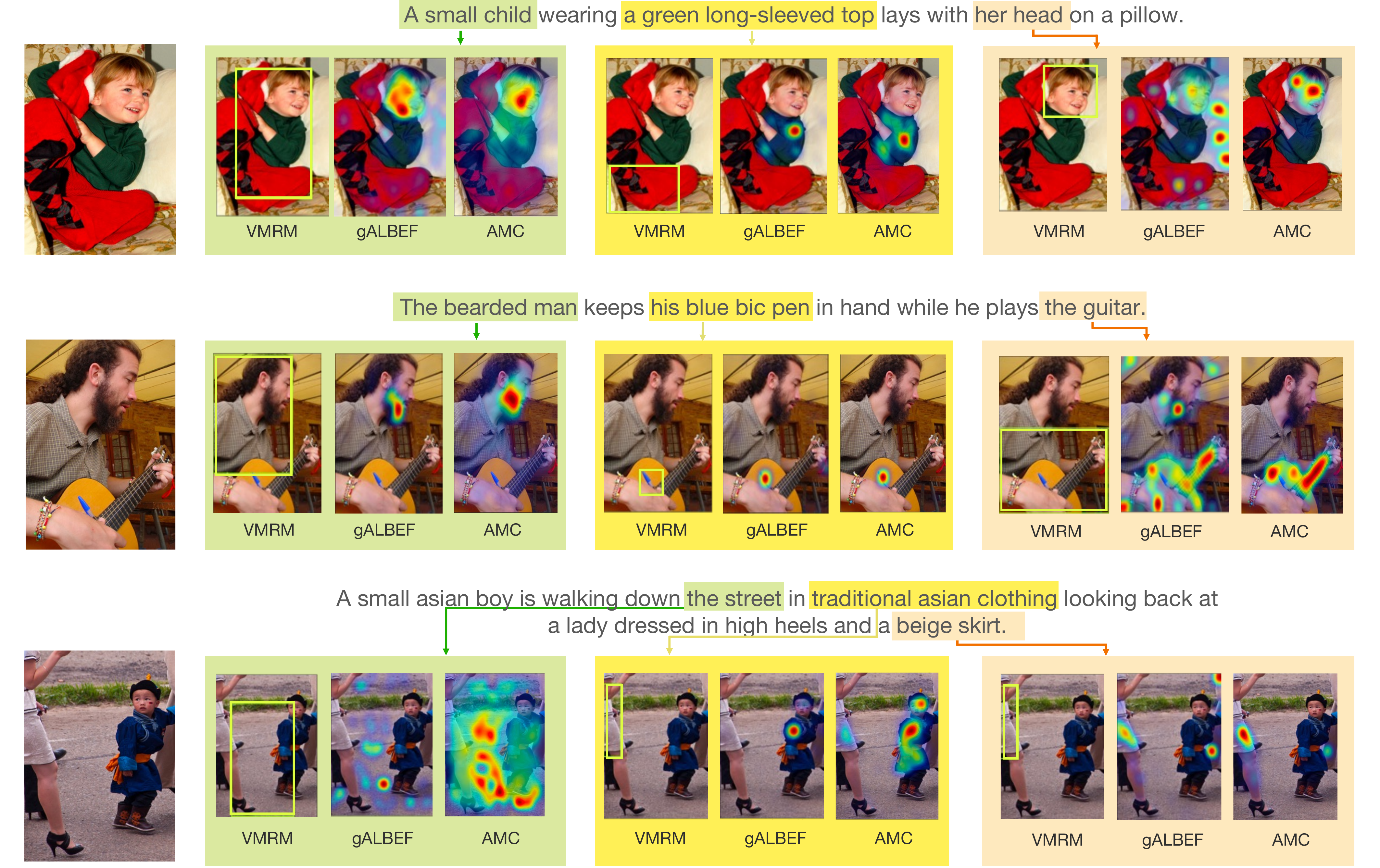}
\caption{Qualitative comparison of the generated explanations for various images and input phrases. First column: original images from Flickr30k Entities; in each colored area from left to right: bounding boxes selected by VMRM; heatmaps generated by gALBEF; heatmaps generated by our method. On the top of each group of images, we show the caption and target phrases.}
\label{fig:qual_flickr}
\vspace{-0.2in}
\end{center}
\end{figure*}

In Figure \ref{fig:qual_flickr} we show and compare visual explanations obtained by our model against those obtained by VMRM~\cite{dou2021improving} and GradCAM heatmaps generated by gALBEF. The text input for VMRM is the whole caption and a phrase and it produces a bounding box prediction. The model locates the positions of the phrase in the caption and selects boxes corresponding to the phrase with context information. For gALBEF and our method, the text input is only a phrase. We found our method can get more accurate and more complete objects from phrases. For example, in the last row, our method can provide a more precise heatmap for the referred object for {\em traditional asian clothing} than gALBEF, and in this case, VMRM is confused by the clothing from the woman instead of the boy. Additionally, in the second row, when the model is asked to find ``the guitar'', our method can accurately cover the guitar, but gALBEF covers several unrelated regions that probably contribute to the detection of \texttt{guitar} but do not provide an explanation that aligns with what a human would annotate for this image. We provide more qualitative results in our supplementary material.

\begin{table}[t!]
\small
\centering
\setlength\tabcolsep{3pt}
\renewcommand{\arraystretch}{1.2}
\begin{tabular}{ l c c c c cc }
\toprule
\multirow{2}{*}{\bf Data} && \multirow{2}{*}{\bf Flickr30k} & \multirow{2}{*}{\bf ReferIt} && \multicolumn{2}{c}{\bf RefCOCO+} \\
\cmidrule{6-7}
 && & && test A & test B\\
\midrule
object boxes && 86.49 & 71.51 && 78.89 & 61.16\\
region boxes && 85.14 & 68.01 && 77.89 & 61.26\\
\midrule
both && 86.59 & 73.17 && 80.34 & 64.55\\
\bottomrule
\end{tabular}
\caption{We conduct an ablation study to evaluate the effect of using "box" annotations corresponding to box + object labels + object attributes from Visual Genome, and "region" annotations corresponding to region boxes + region descriptions from Visual Genome.}
\label{ab_data}
\end{table}

\begin{table}[t!]
\centering
\setlength\tabcolsep{3pt}
\renewcommand{\arraystretch}{1.2}
\begin{tabular}{ l c c c c cc }
\toprule
\multirow{2}{*}{\bf Loss} && \multirow{2}{*}{\bf Flickr30k} & \multirow{2}{*}{\bf ReferIt} && \multicolumn{2}{c}{\bf RefCOCO+} \\
\cmidrule{6-7}
 && & && test A & test B\\
\midrule
$\mathcal{L}_{\mathrm{cosine}}$ && 84.85 & 70.56 && 76.41 & 60.81\\
$\mathcal{L}_{\mathrm{mean}}$ && 82.83 & 67.65 && 75.34 & 56.90\\
$\mathcal{L}_{\max}$ && 86.56 & 72.47 && 80.34 & 64.47\\
\midrule
$\mathcal{L}_{\mathrm{amc}}$ && 86.59 & 73.17 && 80.34 & 64.55\\
\bottomrule
\end{tabular}
\caption{We conduct an ablation study to evaluate the contribution of $\mathcal{L}_{\mathrm{max}}$ and $\mathcal{L}_{\mathrm{mean}}$ to our final accuracy and an alternative loss based on cosine similarities $\mathcal{L}_{\mathrm{cosine}}$.}
\label{ab_loss}
\end{table}

\subsection{Ablation Studies}

In this section, we present ablations against several choices in our model and contributing factors. Particularly we investigate how large is the effect from box supervision from Visual Genome both from object boxes, and region boxes.

\textbf{Box Supervision.} As described in section~\ref{data}, VG~\cite{krishna2017visual} includes regions with descriptions and objects with attributes. We evaluate our method on each separately. For Flickr30k Entities and the ReferIt dataset, boxes with generated descriptions using attributes lead to better results than regions with descriptions. We believe this is caused by accurate localization information provided by boxes and spatial information from box descriptions. For RefCOCO+ test B, regions with descriptions perform slightly better than boxes with attributes. By combining boxes, regions and two kinds of descriptions, we obtain better alignment between phrases and image subareas. Our full set of results from this experiment are in Table~\ref{ab_data}.

\textbf{Loss Choices.}
Instead of calculating our margin loss as in Eq.~\ref{eq:amc}, we calculate and minimize the cosine distance between $M$ and $A$. Therefore, the generated heatmap will be closer to the box mask. Results for all of our choices that we considered in our objective function are presented in Table~\ref{ab_loss}. For all datasets, our method outperforms this cosine distance loss $\mathcal{L}_{\mathrm{cosine}}$, proving our method is a better way to use box information than a perhaps more straightforward dot product optimization. Furthermore, we evaluate two components in~Eq.~\ref{eq:amc}: $\mathcal{L}_{\mathrm{mean}}$ and $\mathcal{L}_{\max}$. We find $\mathcal{L}_{\max}$ is very significant in AMC, but $\mathcal{L}_{\mathrm{mean}}$ also provides complementary information, especially for the ReferIt dataset. In general, combining two terms leads to a more comprehensive grounding ability but using $\mathcal{L}_{\max}$ alone is also very competitive.

\section{Conclusion}
\label{sec:conclusion}
In this paper, we proposed Attention Map Consistency (AMC). From the intuition that a model should focus on meaningful regions guided by location information, we design an objective function that optimizes gradient-based explanation maps. Our approach achieves superior results on visual grounding compared to other methods with a similar level of supervision. It particularly surpasses methods relying on an  object detector pretrained on Visual Genome.

\paragraph{Acknowledgments} This work was supported by gifts from Adobe Research and NSF Award IIS-2201710. We are also thankful for positive comments and suggestions from anonymous reviewers.

{\small
\bibliographystyle{ieee_fullname}
\bibliography{egbib}
}

\include{appendix}

\end{document}

%% file: appendix.tex


\section{Supplementary}

\begin{figure*}[b!]
\centering
\hspace{0.3in}
\includegraphics[width=0.9\textwidth]{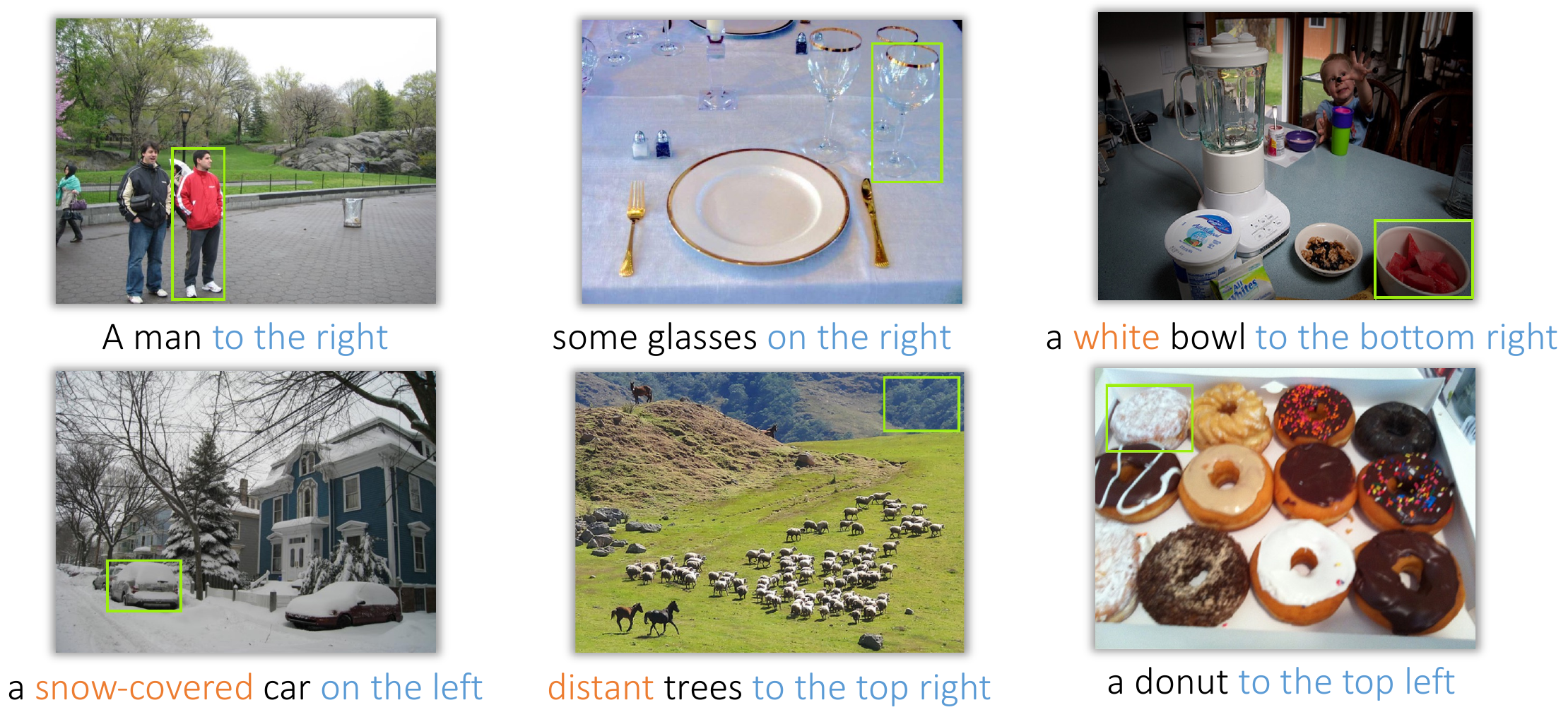}
\caption{We show some constructed textual descriptions with colored attributes and spatial references.}
\label{fig:sup}
\end{figure*}

In this section we provide more details for experiments on RefCOCO+~\cite{yu2016modeling} and ReferIt~\cite{kazemzadeh2014referitgame}, and provide additional qualitative examples.

\begin{table}[h!]
\centering
\setlength\tabcolsep{2.4pt}
\renewcommand{\arraystretch}{1.45}
\begin{tabular}{ l c c c c c cc }
\toprule
\multirow{2}{*}{\bf Method} && \multicolumn{2}{c}{\bf RefCOCO+$\ddagger$}  && \multicolumn{2}{c}{\bf RefCOCO+$\mathsection$} \\
\cmidrule{3-7}
 && test A & test B && test A & test B\\
\midrule
InfoGround~\cite{gupta2020contrastive} && 39.80 & 41.11 && 40.10 & 40.62\\
VMRM~\cite{dou2021improving} && 58.87 & 50.32 && 60.29 & 50.39\\
ALBEF~\cite{li2021align} && 69.37 & 53.77 && 69.40 & 54.04\\
AMC && 80.34 & 64.55 && 80.33 & 65.02\\
\bottomrule
\end{tabular}
\vspace{0.12in}
\caption{We show pointing accuracy results on the \mbox{RefCOCO+} validation and testing sets with original and clean images. $\ddagger$ indicates original image splits and $\mathsection$ indicates splits with clean images.}
\label{supp:clean_data}
\end{table}

\textbf{RefCOCO+ Clean} As discussed in Anderson~et~al~\cite{anderson2018bottom}, there are around $51\mathrm{K}$ images from Visual Genome (VG)~\cite{krishna2017visual} that are also present in the COCO dataset~\cite{lin2014microsoft}. Moreover, images in the RefCOCO+ validation/testing sets come from the COCO dataset as well. While there is no overlap in the training, validation and testing sets for RefCOCO+, methods that use VG to pretrain object detectors might use some overlapping data which would make object detectors on some part of the validation and testing sets artificially accurate. In order to fully investigate whether this issue affects the generalization of previous methods, we further explore a more restricted version of the validation and test sets for RefCOCO+ so that no overlap exists with VG and re-run previous methods along with our method on this subset. After cross-referencing images in the VG training set from Anderson~et~al~\cite{anderson2018bottom} and images from the RefCOCO+ validation/testing sets, we find $574$ and $569$ overlapping images in the RefCOCO+ validation/testing sets. 
In order to correct this, we also evaluate and compare our method with previous methods on a {\em clean} version of the RefCOCO+ validation and testing sets with $926$ and $931$ images respectively. 

Table \ref{supp:clean_data} shows that in fact this overlap did not have much of an effect on previous methods -- and our method also performs at a high accuracy. Our method still outperforms VMRM~\cite{dou2021improving} and InfoGround~\cite{gupta2020contrastive} by a large margin. We also report results for ALBEF~\cite{li2021align} and compare it with InfoGround and VMRM which uses bounding boxes for object detectors during training. Even though ALBEF does not use any box information, it still achieves good performance on the RefCOCO+ dataset. Our method, which uses box information, can further improve the pointing accuracy results under both settings.

\textbf{Sample Spatial Prompts}\: In our main paper, we discuss how to construct textual descriptions using bounding boxes and attributes. In Figure~\ref{fig:sup}, we show several examples of such constructed data. In total, we generate $924,807$ text descriptions using attributes and $168,442$ descriptions with spatial references.

\begin{figure*}[t!]
\begin{center}
\includegraphics[width=0.9\textwidth]{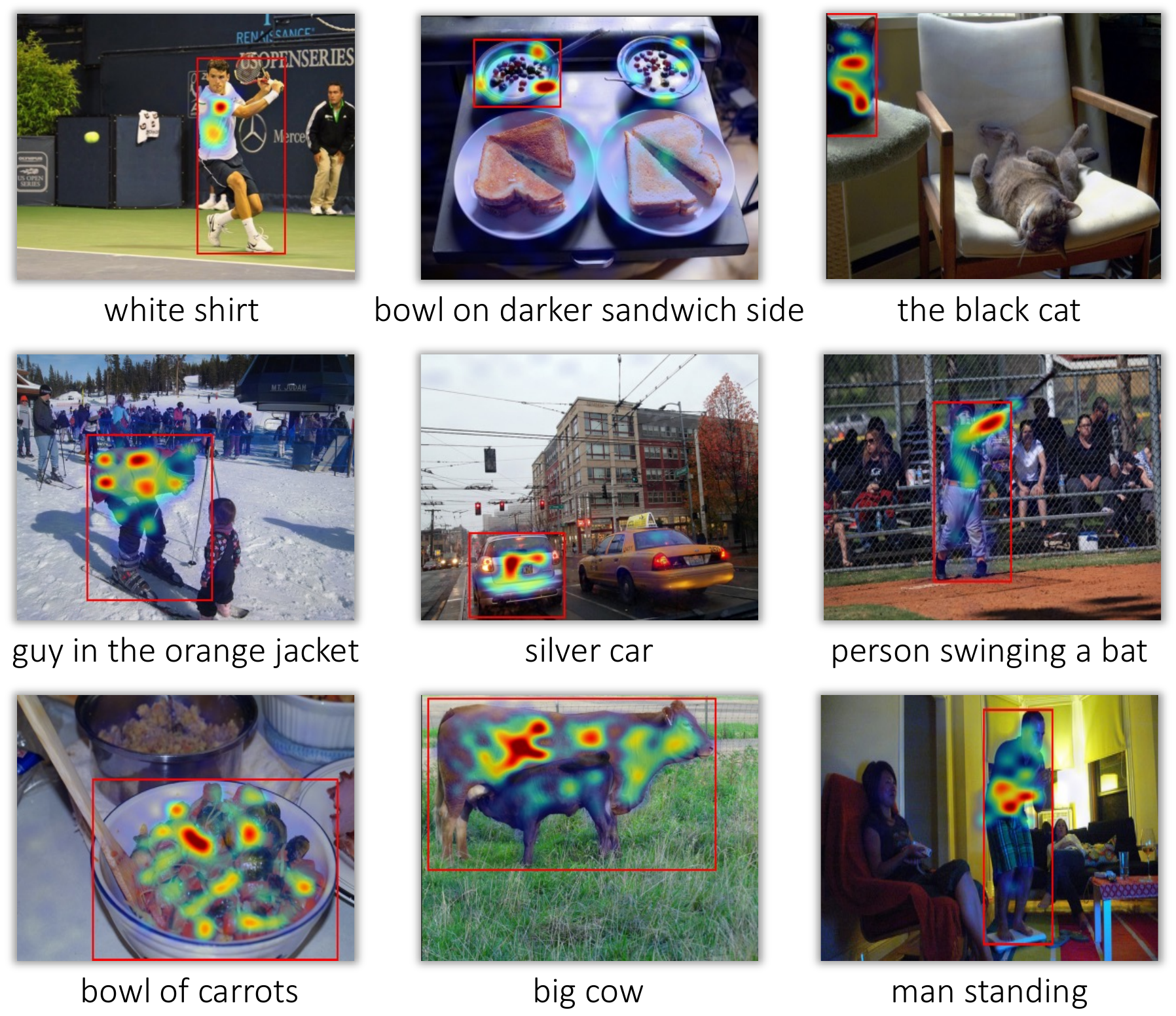}
\hspace{-0.3in}
\caption{We show more qualitative examples for the RefCOCO+ testing set. Ground truth boxes are marked as red boxes. Below each image we provide with one input phrase.}
\label{fig:sup_refcoco}
\end{center}
\end{figure*}

\begin{figure*}[t!]
\begin{center}
\includegraphics[width=0.9\textwidth]{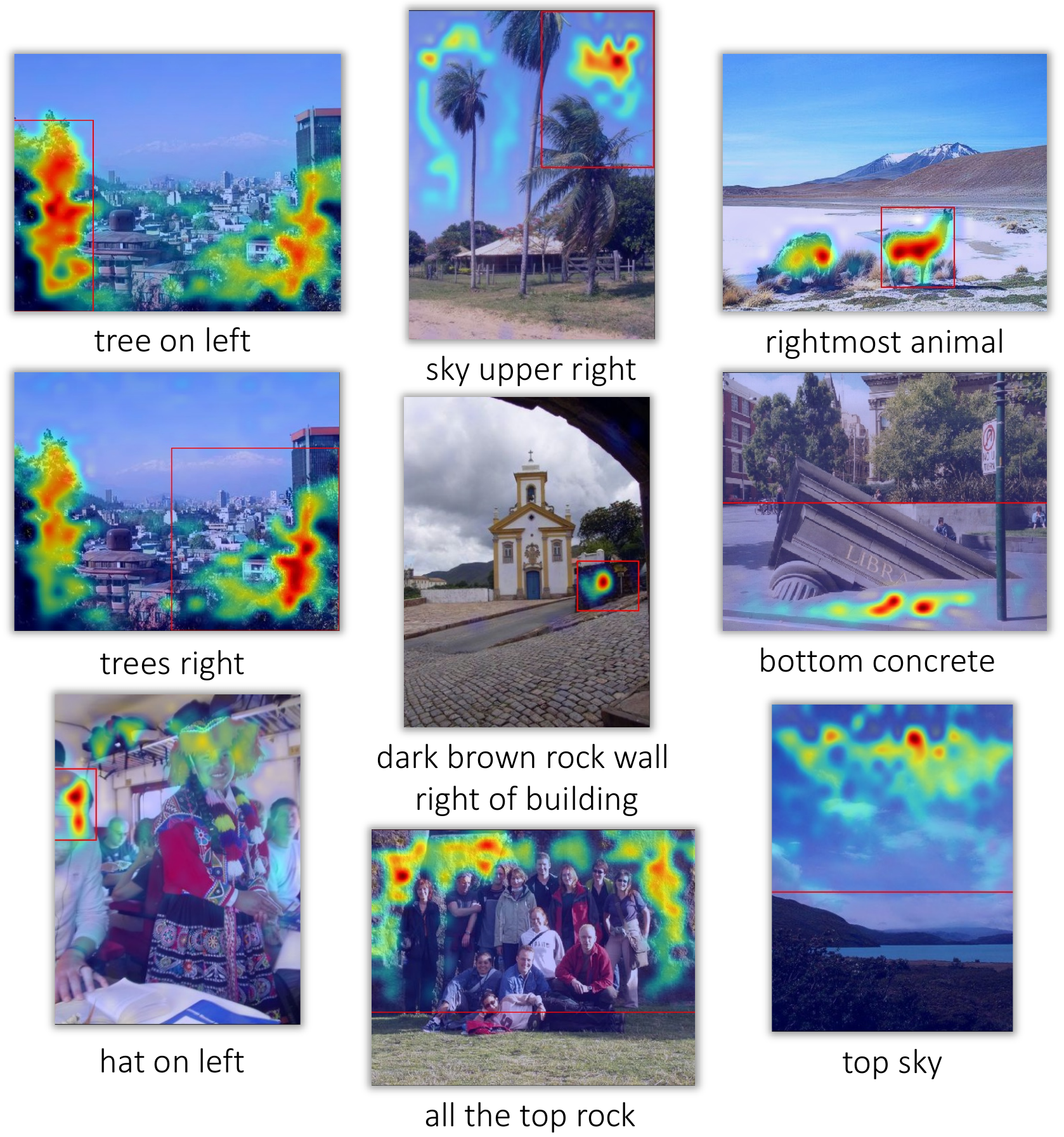}
\hspace{-0.3in}
\caption{We show more qualitative examples for the ReferIt testing set. Ground truth boxes are marked as red boxes. Below each image we provide with one input phrase.}
\label{fig:sup_refclef}
\end{center}
\end{figure*}

\textbf{Qualitative Results}\: Figure~\ref{fig:sup_refcoco} shows additional qualitative results on the test set of RefCOCO+ and Figure~\ref{fig:sup_refclef} shows similar qualitative results for ReferIt. We show heatmaps generated by our method given images and text phrases. Our model can successfully localize the target object even though there exist other similar objects in the same image.



